\documentclass[runningheads]{llncs}
\usepackage{float}
\usepackage{makecell}
\usepackage{xcolor}
\usepackage{booktabs}
\usepackage{tabularx}
\usepackage{multirow}
\usepackage[T1]{fontenc}
\usepackage{graphicx,verbatim}
\usepackage{hyperref}
\begin{document}
\title{Traumatic Brain Injury Segmentation using an Ensemble of Encoder-decoder Models}
\titlerunning{TBI segmentation using Ensemble of Enc-Dec Models}

\author{Ghanshyam Dhamat\inst{1} \and Vaanathi Sundaresan\inst{1,2}}  
\authorrunning{G. Dhamat et al }
\institute{Department of Computational and Data Sciences, Indian Institute of Science
Bengaluru, Karnataka 560012, India  \\ \and
Corresponding author:
\email{vaanathi@iisc.ac.in}}

\maketitle            
\begin{abstract}
The identification and segmentation of moderate-severe traumatic brain injury (TBI) lesions pose a significant challenge in neuroimaging. This difficulty arises from the extreme heterogeneity of these lesions, which vary in size, number, and laterality, thereby complicating downstream image processing tasks such as image registration and brain parcellation, reducing the analytical accuracy. Thus, developing methods for highly accurate segmentation of TBI lesions is essential for reliable neuroimaging analysis. This study aims to develop an effective automated segmentation pipeline to automatically detect and segment TBI lesions in T1-weighted MRI scans. We evaluate multiple approaches to achieve accurate segmentation of the TBI lesions. The core of our pipeline leverages various architectures within the nnUNet framework for initial segmentation, complemented by post-processing strategies to enhance evaluation metrics. Our final submission to the challenge achieved an accuracy of 0.8451, Dice score values of 0.4711 and 0.8514 for images with and without visible lesions, respectively, with an overall Dice score of 0.5973, ranking among the top-6 methods in the AIMS-TBI 2025 challenge. The Python implementation of our pipeline is publicly available.

\end{abstract}

\section{Introduction}
\begin{figure}[H]
    \centering
    \includegraphics[width=0.85\linewidth]{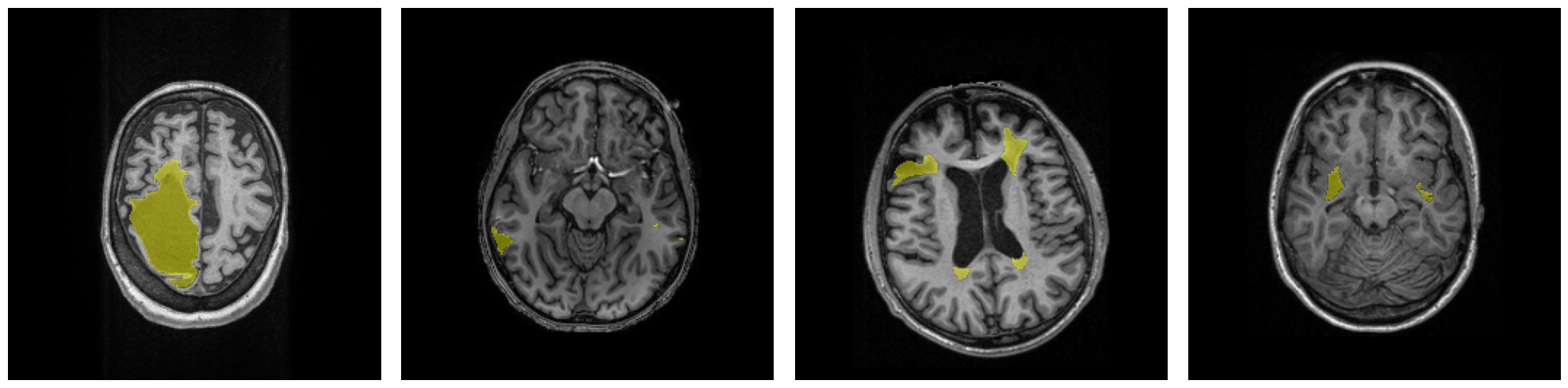}
    \caption{Heterogeneity in the appearance of traumatic brain injury lesions on T1-weighted MRI (images overlaid with manual annotations of TBI lesions in yellow). }
    \label{fig: Fig 1}
\end{figure}

Traumatic brain injuries (TBI) occur when external forces cause brain movement within the skull, resulting in biological and functional alterations \cite{aims-tbi}. These injuries lead to heterogeneous lesions, a key characteristic of moderate-severe (MS)-TBI. These lesions can be focal or diffuse, varying in size and number, affecting different brain tissues, and potentially appearing in similar regions on both sides. Fig.~\ref{fig: Fig 1} illustrates the heterogeneity in MS-TBI lesion characteristics across different subjects. If these lesions are overlooked during image processing tasks like registration or brain parcellation, it can lead to significant errors in analytical outcomes \cite{aims-tbi}. 
Although manual segmentation of MS-TBI lesions is considered the gold standard, it is time-consuming, subjective, and shows high variability between raters. This drives research towards automated methods for identifying and delineating lesions. Automated segmentation speeds up the process and can provide additional information to clinicians. 

The AIMS-TBI 2025 challenge aims to develop automated algorithms that effectively detect and segment lesions while addressing the heterogeneous nature of MS-TBI lesions. The challenge focuses on T1-weighted MRI as they are most commonly used by the ENIGMA TBI consortium and show less parameter variation compared to other MRI modalities \cite{aims-tbi}.\\
In this work, we investigate various approaches, including ensemble networks, based on the nnUNet framework to train segmentation models. These approaches incorporate variations in encoder-decoder architectures and classifier-based post-processing strategies to enhance accuracy. We also experiment with radiomics-based features for improving TBI lesion segmentation. Evaluation on heterogeneous AIMS-TBI challenges data shows the effectiveness of ensemble models in accurate TBI lesion segmentation.

\section{Method}
We explore automated pipelines to identify and segment TBI lesions. The process begins with preprocessing applied to all volumes. We investigate settings of the encoder-decoder model and their ensemble, combined with a radiomics feature-based classifier. The components are detailed below.\\

\hspace{-1.5em} \textbf{Data preprocessing:} \label{ssec:preproc}
Before training, we apply bias field correction using ANTsPyx library in python \cite{ANTSx} to mitigate low-frequency intensity inhomogeneities in the MRI volumes. 

\subsection{Traumatic brain injury initial candidate segmentation} 

We build the pipelines based on the widely used nnUNet framework \cite{nnunet_github,nnunetv2}, with configurations to achieve optimal performance while handling all preliminary steps involving resampling, normalization, and tight cropping. \\

\textbf{UNET (integrated with nnUNet framework).} We initially use a 3D UNet \cite{UNET} integrated within the nnUNet framework. The UNet architecture is a convolutional neural network with an encoder-decoder structure, interconnected with skip connections which combine texture (from encoder) with localization features (in the decoder). Each stage consists of double convolution operations, enhancing the network's ability to learn hierarchical features for segmentation. \\

\begin{figure}[H]
    \centering
    \begin{minipage}{0.4\linewidth}
        \centering
        \raisebox{-0.5\height}{\includegraphics[width=\linewidth]{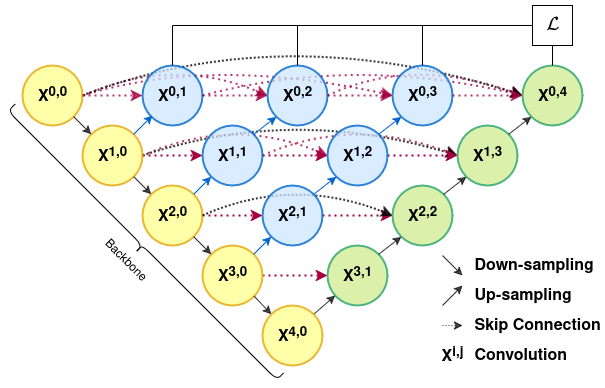}}
        \\(a)
    \end{minipage}
    \quad
    \begin{minipage}{0.4\linewidth}
        \centering
        \raisebox{-0.2\height}{\includegraphics[width=\linewidth]{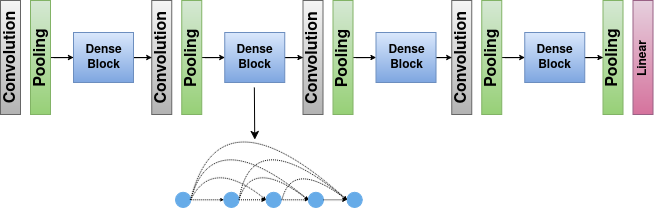}}
        \\[1em](b)
    \end{minipage}
    
    \caption{(a) UNET++ architecture with nested skip connections and (b) DenseNet architecture with dense block where each layer uses feature map from preceding layers.}
    \label{fig:unet-densenet}
\end{figure}

\textbf{UNETPP (integrated with nnUNET framework).} As an improvisation to the vanilla UNet, the UNet++ architecture \cite{unetpp} integrates nested, dense skip pathways, as shown in Fig.~\ref{fig:unet-densenet}, designed to capture finer-grained details in complex structures. 

\subsection{2D classification of slices with visible lesions vs without lesions }  
\label{ssec:2dclass}
As a post-processing strategy for removal of false positives (FPs), we classify slices with segmented regions as slices with visible lesions vs without lesion using DenseNet architecture \cite{Densenet} on 2D slices from 3D volumes. The DenseNet architecture contains dense blocks within which each layer connects to every subsequent layer, facilitating feature reuse and improving information flow. These dense blocks are placed between transition layers that change feature map size. Classification was performed only on slices where the segmentation model predicted a potential lesion. To cut down on minor false positives in cases without a lesion, we restricted the use of 2D classifiers to subjects whose segmented lesion volume was less than 2000 mm$^{3}$.
\subsection{Radiomics-based classifier for refining segmentation} 
\label{ssec:radiomics}
To further refine the segmentation model's output, we use a radiomics-based classifier to remove false positive voxels. We extract voxel-based radiomics features capturing intensity, shape, and texture information using Python's PyRadiomics library \cite{radiomics}. Among various classifiers explored (using Scikit-learn package), including SVM \cite{SVM} and Random forest {\cite{randomforest}}, XGBoost \cite{XGBoost} is selected since it offered 71\% accuracy (outperformed others by at least 11\%). Due to computational constraints, we limit the processing to lesions with a volume $<$ 1000 voxels and use the top 25 important features for training XGBoost.\\

The various components explained above are utilized into ensembles explained as different settings in experiments (Sec.~\ref{ssec:exps}).
\section{Experimental Setup}
\subsection{Dataset Overview}
The AIMS-TBI 2025 challenge dataset \cite{aims-tbi} includes 875 T1-weighted MRI scans, divided into 552 training, 100 validation, and 223 unseen test cases, with balanced distribution across sets. The scans were collected from 13 sites using 1.5T and 3T scanners. The dataset includes patients aged 5-85 years, with 64\% male subjects. Lesions were segmented through a three-step process for high-quality ground truth annotations. All MRI images were pre-processed by defacing to protect patient privacy \cite{aims-tbi}. Of the 552 training samples available, we kept 24 cases for testing and used the remainder for training and validation (with training:validation=90:10). Among the 24 samples, 5 were from subjects with no visible lesions, and the rest were from those with lesions. Training phase results are reported on the 24 testing samples, while the final evaluation phase results are reported on 223 unseen test cases, after submission for an in-house evaluation by organizers.
\subsection{Experiments}
\label{ssec:exps}
We explore the following settings trained using the parameters specified in Sec.~\ref{ssec:imp_details}: \textbf{Setting 1: (Baseline U-Net):}
The baseline 3D UNet is trained using the default nnUNet pipeline with 5-fold cross-validation on the 528 training cases, and tested locally on 24 held-out samples. We generate predictions by averaging the probability maps from 5 folds. \textbf{Setting 2 (UNETPP):}
We trained the UNET++ architecture integrated within the nnUNet framework on 528 training samples. \textbf{Setting 3 (UNET + slice-based classifier):} To reduce false positives, 2D slice-based visible lesion vs no lesion classification is trained using DenseNet (sec.~\ref{ssec:2dclass}). To mitigate class imbalance, we exclude the first 45 slices from each volume in the axial direction (starting from the neck) where lesions were negligible, and ensure an equal number of positive and negative slices per batch. The model was trained on slices from 390 volumes and validated on 80. If the classifier predicted ``no visible lesion'' on more than 50\% of the slices containing segmentation, those predictions were removed. 
\textbf{Setting 4 (UNETPP + slice-based classifier):} This setting combined the UNET++ segmentation model with the same DenseNet-based classification approach from Setting 3. 
\textbf{Setting 5 (UNET + slice-based classifier + radiomics classifier):}
To evaluate the cumulative effect of postprocessing steps on segmentation accuracy and FP reduction, in this setting, the UNET segmentation was first refined by the slice-based classifier and then further refined at the voxel level by the radiomics-based XGBoost classifier (sec.~\ref{ssec:radiomics}). 
\textbf{Setting 6 (UNETPP + slice-based classifier + radiomics classifier):}
Similar to setting 5, this setting is built upon setting 4 by adding a second stage of voxel-level radiomics-based XGBoost classifier.
\textbf{Setting 7 (UNET + UNETPP):} We evaluate an ensemble approach to leverage the strengths of both UNet and UNET++ by averaging the probability maps from the 5-fold UNet ensemble (Setting 1) and the single-fold UNET++ model (Setting 2).
\subsection{Implementation Details} 
\label{ssec:imp_details}
Experiments are conducted on an Intel i9-10980XE CPU, 128 GB RAM, and two NVIDIA RTX A6000 GPUs (each with 48 GB memory) using Python 3.10 and PyTorch 2.8. For training segmentation models, we use a combination of Dice loss and cross-entropy loss. We trained for 1000 epochs, with a batch size of 2, an initial learning rate of $1e^{-2}$ with a linear decay scheduler. For nnUNet training, we use a patch size of 128 $\times$ 160 $\times$ 112 voxels. Data augmentations include flipping, intensity rescaling ($\gamma \in (0.7,1.5)$) and rotation ($\theta \in (-180^{o},180^{o}$)). 

\subsection{Performance metrics and statistical evaluation}
\label{ssec:perf_eval}
We use Dice Similarity Coefficient (DSC) and subject-level accuracy for the evaluation of TBI segmentation.
To provide a comprehensive assessment, three DSC metrics were determined as part of the challenge:
mean DSC value calculated only for subjects with visible lesion (\textbf{DSC-Lesion}), for subjects with no visible lesion (\textbf{DSC-no-Lesion}) and finally for all subjects (\textbf{Overall DSC}). To determine the statistical significance of the results, we perform two-tailed paired t-tests between the best settings and the baseline setting 1 in the training phase, and between the settings in the final evaluation phase. 
\section{Results and Discussion}
\begin{figure}[h]
    \centering
    \includegraphics[width=\linewidth]{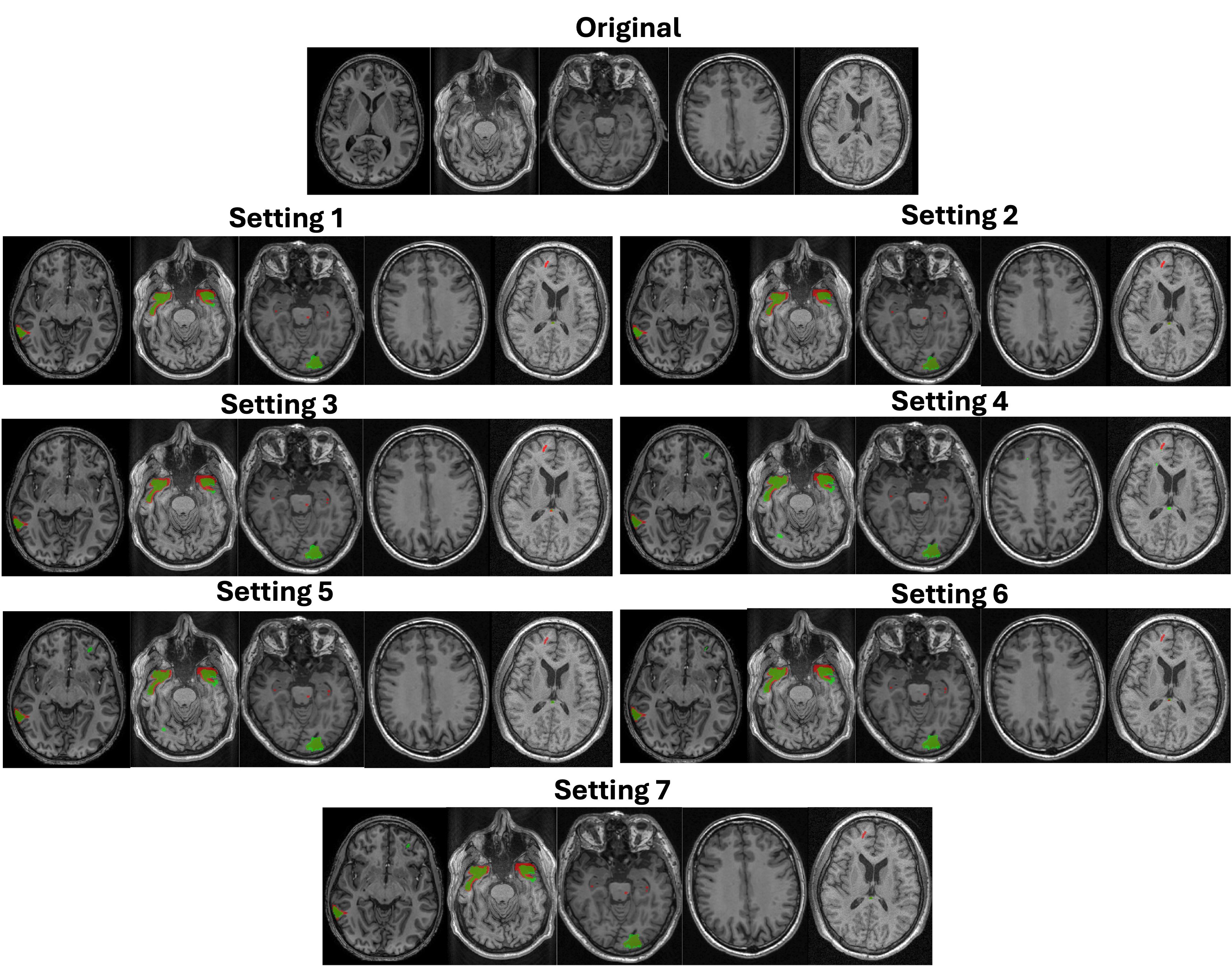}
    \caption{Comparison of results from various ensemble settings, with each setting shown for five sample subjects (left to right). Predicted lesions are shown in green while the ground-truth TBI lesions are shown in red. Results indicate better performance with settings 3 and 7.}
    \label{fig: Fig 3}
\end{figure}

Table~\ref{tab:evaluation_metrics} shows the segmentation performance for our experimental settings, and Fig.~\ref{fig: Fig 3} shows the visual results. 
\begin{table}[htbp]
\caption{Comparison of segmentation performance across various settings, for training and final evaluation phases. Statistical significance is assessed for the best settings with respect to the baseline setting 1 using a two-tailed paired t-test, 
with * indicating significant improvement with p-value < 0.05.}
    \centering
    \small
    \begin{tabularx}{\textwidth}{X c c c c c}
        \toprule
        \textbf{Phase} &
        \textbf{Exp. Setting} & 
        \textbf{Accuracy} & 
        \textbf{DSC-Lesion} &
        \textbf{DSC-no-Lesion} &
        \textbf{Overall DSC} \\
        \midrule
        \multirow{7}{*}{\parbox[c]{1.7cm}{Training phase: local test data created from the training data}} 
            & Setting 1 & 1.00 & $0.70 \pm 0.11$ & $1.00 \pm 0.0$ & $0.77 \pm 0.16$ \\
            & Setting 2 & 0.916 & $0.72 \pm 0.095$ & $0.60 \pm 0.48$ & $0.697 \pm 0.24$ \\
            & Setting 3 & 1.00 & $(0.68 \pm 0.12)^{*}$ & $1.00 \pm 0.0$ & $(0.75 \pm 0.17)^{*}$ \\
            & Setting 4 & 1.00 & $0.695 \pm 0.12$ & $1.00 \pm 0.0$ & $0.76 \pm 0.16$ \\
            & Setting 5 & 1.00 & $(0.673 \pm 0.13)^{*}$ & $1.00 \pm 0.0$ & $(0.74 \pm 0.17)^{*}$ \\
            & Setting 6 & 1.00 & $0.687 \pm 0.13$ & $1.00 \pm 0.0$ & $0.75 \pm 0.17$ \\       
            & Setting 7 & 1.00 & $0.698 \pm 0.11$ & $1.00 \pm 0.0$ & $0.76 \pm 0.16$ \\
        \midrule
        \multirow{2}{*}{\parbox[c]{1.7cm}{Final phase: Test data}} 
            & Setting 3 & $0.80$ & $0.41\pm0.34$ & $0.86\pm0.34$  & $0.56\pm0.40$  \\
            & Setting 7 & $(0.84)^{*}$ & $(0.47\pm 0.31)^{*}$ & $0.85\pm0.36$  & $(0.59 \pm 0.37)^{*}$ \\
        \bottomrule
    \end{tabularx}
    \label{tab:evaluation_metrics}
\end{table}
On the local test set, the baseline UNET model (setting 1) established a strong benchmark with a DSC-Lesion of $0.7\pm0.11$. It was particularly reliable in correctly identifying subjects with no visible lesions; however, it tended to merge distinct, adjacent lesions, thus yielding a lower DSC than that of setting 2. Setting 2 with the UNETPP model achieved the best DSC-Lesion ($0.72\pm0.095$), however, it suffered on subjects with no visible lesion. This indicates that UNETPP, designed to capture finer details, was overly sensitive to image noise, which resulted in small FPs as shown in Fig.~\ref{fig: Fig 3} (also setting 4 involving UNETPP). The setting 7 successfully balanced these characteristics, showcased in Fig. \ref{fig: Fig 3} setting 7 for columns 2 and 4. Overall, setting 7 maintains competitive scores in both cases with and without a lesion. The introduction of post-processing stages yielded mixed results - while the 2D slice-based classifier proved effective in increasing DSC on subjects with no visible lesion, it led to a reduction in DSC for subjects with lesions in both setting 3 and setting 4, with a statistically significant decrease in setting 3 $(0.68 \pm 0.12, p = 0.003)$. DenseNet architecture was opted for the classifier due to its parameter efficiency and strong feature propagation. By conditionally applying this classifier only to cases with small predicted volumes, we created a fast and effective filter against minor FPs without affecting larger, more confident segmentations. We explored the radiomics-based classifier to refine the segmentation by removing FPs at the voxel level. However, from table \ref{tab:evaluation_metrics}, we can see that the DSC-lesion values reduced for setting 6 $(0.687 \pm 0.13)$ and significantly for setting 5 $(0.673 \pm 0.13, p<0.0001)$. The reduction in performance suggests that radiomics features could not sufficiently discriminate lesions from FPs, especially for small TBI lesions. For instance, any FPs likely arose from image artifacts whose low-level texture and shape features, captured by radiomics, were indistinguishable from those of small, true lesions. Furthermore, the computational constraint limited the application of radiomics to lesions $<$ 1000 voxels, and hence, the classifier could not leverage potentially more discriminative features of FPs from larger lesions.
For the final challenge submission, we selected two of our models: the robust UNET + 2D classifier pipeline (setting 3) and UNET + UNETPP ensemble (setting 7). The former represented a stable, well-vetted approach, while the latter was chosen to leverage the diverse feature extraction capabilities of two distinct architectures. On the unseen test data, the ensemble approach (UNET + UNETPP) proved superior, outperforming the post-processing pipeline in setting 3 across all key metrics, achieving a significantly higher overall Dice of 0.59 compared to 0.56 (p-value = 0.013) and an increased accuracy from 0.8 to 0.84 (p-value = 0.004). 
\begin{figure}[H]
    \centering    \includegraphics[width=\linewidth]{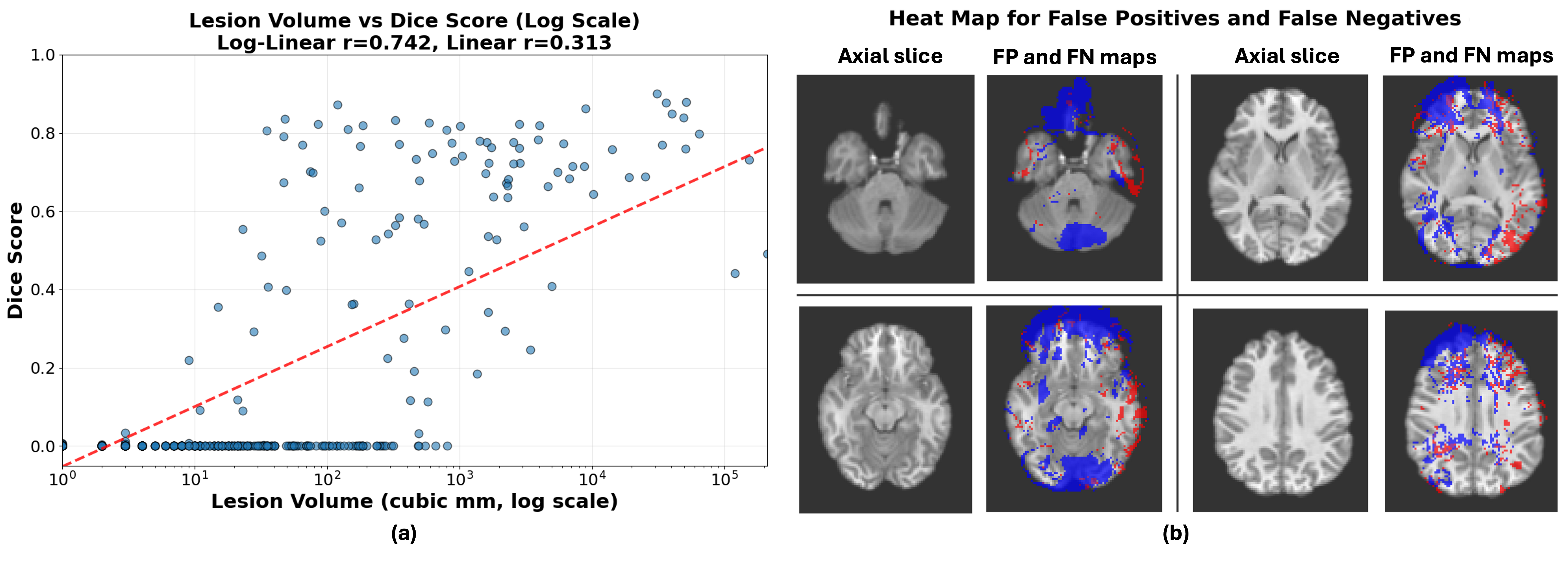}
    \caption{(a) Relationship of lesion size (cubic mm) in log scale vs Dice score on local test set, (b) Heatmap for false positive (red) and false negative (blue) in local test set.}
    \label{fig:fig4}
\end{figure}
Fig.~\ref{fig:fig4} shows that, while our model is mainly overlooking small lesions with volume less than 100 mm$^{3}$, a few large ones are also getting missed due to the lack of representative training instances. The log-linear Pearson coefficient of 0.734 shows an overall positive correlation of the Dice score with lesion size.

The future directions include: (1) use of transformer-based architecture, as they have been shown to provide better results across various tasks due to their ability to model long-range dependencies resulting in more global understanding of the image context, (2) self-supervised pretraining of the segmentation model on a large corpus of unlabeled brain MRIs to learn richer and more robust representations and fine-tuning them to improve segmentation results.
\section{Conclusions}
In this study, we compare various ensemble settings of encoder-decoder architectures to detect and segment MS-TBI lesions on the AIMS-TBI Challenge dataset. Our pipelines are built based on the nnUNet framework, and we also investigate the effect of a classification-based post-processing strategy and the ensemble of networks. Our results on the final evaluation indicate that the ensemble models provided the best segmentation performance with an overall DSC of 0.59, ranking among the top-6 methods in the AIMS-TBI 2025 challenge, including subjects with and without lesion, which is significantly better than other settings. While the postprocessing strategy involving 2D slice-based classification eliminated false positives, there is still scope for improvement in terms of accurate identification of small TBI lesions. Future directions include exploring transformer-based architectures and self-supervised preprocessing techniques for better performance. The python code for our pipelines has been made available at \url{https://github.com/ghanshyamdhamat/aims\_tbi\_2025.git}.
\paragraph{\textbf{Acknowledgements. }}This work was supported by DBT Wellcome Trust India Alliance Fellowship [IA/E/22/1/506763], the Council of Scientific \& Industrial Research (CSIR) under its ASPIRE program [25WS(013)/2023-24/EMR-II/ASPIRE], the Science and Engineering Research Board Start-up Research Grant [SRG/2023/001406], Siemens Healthineers-CDS Collaborative Laboratory of Artificial Intelligence in Precision Medicine, India and Pratiksha Trust, Bangalore, India [FG/PTCH-23-1004].

\paragraph{\textbf{Disclosure of Interests. }}The authors have no competing interests to declare that
are relevant to the content of this article.
\bibliographystyle{splncs04}
\bibliography{reference}

\end{document}